\documentclass[runningheads]{llncs}
\usepackage[T1]{fontenc}
\usepackage{graphicx}
\newcommand{\E}[2]{ {\rm E}_{#1} \left[ #2 \right] }
\newcommand{\vbar}{\,|\, }

\newcommand{\argmin}[1]{ \underset{#1}{\operatorname{arg\,min}} }

\usepackage{amsmath}
\usepackage{amssymb}
\usepackage{amsfonts}
\usepackage{microtype}
\usepackage{graphicx}
\usepackage{booktabs}
\usepackage{mathtools}
\usepackage{bm}
\usepackage{multirow}
\usepackage{makecell}
\usepackage{graphicx}
\usepackage[belowskip=-10pt,aboveskip=0pt]{caption}
\usepackage{subcaption}
\usepackage[misc]{ifsym}
\usepackage{algorithm}
\usepackage{algpseudocode}

\begin{document}
\title{Sparse Horseshoe Estimation via Expectation-Maximisation}
%
%
\author{Shu Yu Tew\inst{1}\orcidID{0000-0002-6560-3131} \and
Daniel F. Schmidt\inst{1}(\Letter)\orcidID{0000-0002-1788-2375} \and
Enes Makalic\inst{2}\orcidID{0000-0003-3017-0871}}
\authorrunning{S.Tew et al.}
%
\institute{Department of Data Science and AI, Monash University, Clayton, Australia
\email{ 
\{shu.tew, daniel.schmidt\}@monash.edu}\\ \and
Centre for Epidemiology and Biostatistics, The University of Melbourne, Australia\\
\email{emakalic@unimelb.edu.au}
}

\toctitle{Sparse Horseshoe Estimation via Expectation-Maximisation}
\tocauthor{Shu~Tew,Daniel~F~Schmidt,Enes~Makalic}

\maketitle              
\begin{abstract}
The horseshoe prior is known to possess many desirable properties for Bayesian estimation of sparse parameter vectors, yet its density function lacks an analytic form. As such, it is challenging to find a closed-form solution for the posterior mode. Conventional horseshoe estimators use the posterior mean to estimate the parameters, but these estimates are not sparse. We propose a novel expectation-maximisation (EM) procedure for computing the MAP estimates of the parameters in the case of the standard linear model. A particular strength of our approach is that the M-step depends only on the form of the prior and it is independent of the form of the likelihood. We introduce several simple modifications of this EM procedure that allow for straightforward extension to generalised linear models. In experiments performed on simulated and real data, our approach performs comparable, or superior to, state-of-the-art sparse estimation methods in terms of statistical performance and computational cost.

\end{abstract}
\section{Introduction}\label{introduction}

Sparse modelling has become increasingly important in statistical learning due to the growing demand for the analysis of high dimensional data with parameters whose dimension exceeds the sample size. Inference under such a setting often involves the assumption that the parameter vector is sparse. This implies that most of the components in the vector are insignificant and should be removed to achieve a good estimate. The lasso (i.e., $\ell_1$-penalised regression)~\cite{tibshirani1996regression} is a popular sparsity inducing technique well-known for its ability to simultaneously perform variable selection and coefficient estimation. The lasso is a convex penalisation technique, and is known to suffer from over-shrinkage and consistency issues~\cite{casella2010penalized}; to combat these, authors introduced non-convex penalty to replace the usual $\ell_1$-norm used in the lasso. Well-known non-convex penalties include the smoothly clipped absolute deviation (SCAD)~\cite{fan2001variable} and the minimax concave penalty (MCP)~\cite{zhang2010nearly}. \\

However, these approaches are all frequentist in nature, usually relying on an additional principle such as cross-validation to select the degree of regularisation, and do not have immediate access to reliable measures of uncertainty to complement the point estimates they produce~\cite{casella2010penalized}. Bayesian inference, on the other hand, naturally quantifies uncertainty directly through the posterior distribution. Bayesian penalised regression offers readily available uncertainty estimates, automatic estimation of the penalty parameters, and more flexibility in terms of penalties that can be considered~\cite{van2019shrinkage}. In this paper, we consider the standard Gaussian linear regression model
\begin{equation} \label{eq:1}
   {\bf y} = \beta_0 {\bf 1}_n + {\bf X}\bm{\beta}  + \bm{\epsilon}, \quad \bm{\epsilon} \sim N({\bf 0 },\sigma^2 {\bf I}_n )
\end{equation}
where ${\bf y} = (y_1, \cdots, y_n)^T \in \mathbb{R}^n$ is a vector of outcome variable, ${\bf X} \in \mathbb{R}^{n \times p}$ is a matrix of predictor variables, $\beta_0 \in \mathbb{R}$ is the intercept, and $\bm{\epsilon} $ is a vector of i.i.d. normally distributed random error with mean zero and unknown variance $\sigma^2$. Given ${\bf y}$ and ${\bf X}$, our goal is to accurately identify and estimate the non-zero components of the unknown regression coefficients $\bm{\beta}=(\beta_1, \cdots, \beta_p) \in \mathbb{R}^p$.

\subsection{Bayesian Penalised Regression}

Most penalised regression methods can be solved in the Bayesian framework by interpreting the estimates of $\bm{\beta}$ as the posterior point estimate of choice under an appropriate prior distribution~\cite{van2019shrinkage}. For example, the lasso estimate can be interpreted as posterior mode estimates when the regression parameters $\bm{\beta}$ follow an independent and identical double exponential (Laplace) prior~\cite{tibshirani1996regression}. Motivated by this relationship, much work has been done in proposing different Bayesian approaches to the sparse estimation problem via a variety of sparsity inducing priors. Some popular examples include the Bayesian Lasso~\cite{park2008bayesian}, the horseshoe estimator~\cite{carvalho2010horseshoe} and the normal-gamma estimator~\cite{brown2010inference}. These estimators are frequently implemented using Monte Carlo Markov Chain (MCMC) algorithms to sample from the posterior distribution, and the resulting Bayesian estimates are usually  posterior means or medians.\\ 

Interestingly, however, while these priors may be sparsity promoting, in the sense that they concentrate probability mass around sparse coefficient vectors, the resulting posterior means or medians will never themselves be sparse. Sparse estimates can be achieved by instead considering the posterior mode, i.e., maximum a posteriori (MAP) estimation~\cite{figueiredo2003adaptive,armagan2010bayesian,bhadra2019horseshoe}. Unfortunately, most implementations of Bayesian MAP-based sparse estimators are not flexible in the sense that they only work for linear models when paired with a specified prior. These methods do not easily generalise to other regression models and are difficult to apply to priors that lack an analytic form (e.g., horseshoe prior, Strawderman-Berger prior and normal-Gamma prior). Alternatively, simple thresholding (``sparsification'') rules for the posterior mean or median of $\beta_j$s can be used to produce sparse estimates, but these tend to lack theoretical justification, and inference can be highly sensitive to the choice of the threshold~\cite{li2017variable,carvalho2010horseshoe}. To address these limitations:
\begin{enumerate}
    \item we propose a novel expectation-maximisation (EM) procedure to solve for the {\em exact} posterior mode of a regression model under the horseshoe prior;
    
    \item we introduce several simple modifications of our base EM procedure that allow for straightforward extension to models beyond the usual Gaussian linear model, e.g., generalised linear models.
\end{enumerate}
Experiments on simulated data and real datasets in Section \ref{sec: results} demonstrate that our proposed Bayesian EM sparse horseshoe estimator can give comparable, if not better, performance when compared to state-of-the-art non-convex solvers in terms of statistical performance and computational cost. As far as the authors are aware, this is the first work that explores the posterior mode under the exact representation of the horseshoe prior, instead of using an approximated density function in place of the horseshoe prior.

\section{Background and Related Work}
The EM algorithm~\cite{dempster1977maximum} is one of the most widely used methods for MAP estimation of sparse Bayesian linear models, with~\cite{figueiredo2003adaptive} and~\cite{kiiveri2003bayesian} being classic papers on this approach. A particular strength, in comparison to approximate methods such as variational Bayes~\cite{wand2011mean}, is that it is guaranteed to converge to exact posterior modes (a stationary point in the likelihood) whenever it can be applied~\cite{dempster1977maximum}. By introducing appropriate latent variables, the hierarchical decomposition of the Laplace prior allows~\cite{figueiredo2003adaptive} for the derivation of an efficient EM algorithm to recover the lasso estimates. Following this work, many authors have attempted similar procedure to achieve sparse estimation using various priors and likelihood models that admit a scale mixture of normals (SMN) representation. This includes the lasso prior~\cite{park2008bayesian}, the generalised double Pareto prior~\cite{armagan2010bayesian} and the horseshoe-like prior~\cite{bhadra2019horseshoe}. \\

In this paper, we focus on the class of global-local shrinkage priors~\cite{polson2010shrink}, and more specifically the horseshoe prior, which has been recognised as a good default prior choice for Bayesian sparse estimation~\cite{bhadra2016default,carvalho2010horseshoe}. The horseshoe prior has a pole at $\beta_j = 0$, and heavy, Cauchy-like tails. These properties are desirable in sparse estimation because they allow small coefficients to be heavily shrunk towards zero while ensuring large coefficients are not over-shrunk. This is in contrast to the popular Bayesian lasso and Bayesian ridge hierarchies that apply the same amount of shrinkage to all coefficients and potentially over-shrink coefficients that are far from zero. The Gaussian linear regression model (\ref{eq:1}) corresponds to the following likelihood  
\begin{eqnarray}\label{eq:likelihood_regression}
    {\bf y}\,|\,{\bf X}, \bm{\beta}, \sigma^2 \; &\sim& \; N_n\left({\bf X}\bm{\beta}, \; \sigma^2\bm{I}_n\right)
\end{eqnarray}
where $N_k(\cdot,\cdot)$ is the $k$-variate Gaussian distribution. In the class of global-local shrinkage priors, each $\beta_j$ is assigned a continuous shrinkage prior centered at $\beta_j = 0$ that can be represented in an SMN form as
\begin{eqnarray}
\nonumber
\beta_j\,|\,\tau^2, \lambda_j^2, \sigma^2 \; &\sim& \; N\left(0, \; \tau^2 \lambda^2_j \sigma^2\right)\\
\nonumber
\lambda_j^2 \; &\sim& \; \pi(\lambda_j^2)d\lambda^2\\
\label{eq:global-local}
\tau^2 \; &\sim& \; \pi(\tau^2)d\tau^2
\end{eqnarray}
where $\tau$ is the {\em global} shrinkage parameter that controls the overall degree of shrinkage, $\lambda_j$ is the {\em local} shrinkage parameter associated with the $j$-th predictor and it controls the shrinkage for individual coefficients, and $\pi(\cdot)$ is an appropriate prior distribution of choice assigned to the shrinkage parameters. Given the (unormalised) joint posterior distribution of the hierarchy (\ref{eq:likelihood_regression})-(\ref{eq:global-local}),
\begin{equation}\label{eq: regression_post}
p(\bm{\beta}, \tau, \bm{\lambda} | {\bf y})\; \propto \; p({\bf y}| \bm{\beta})\cdot \pi(\bm{\beta}|\bm{\lambda},\tau) \cdot \pi(\bm{\lambda}) \cdot \pi(\tau),
\end{equation}
the conventional EM approach to find the posterior mode estimates treats the hyperparameters (i.e., the shrinkage parameters) $\bm{\lambda}$ as ``missing data'' (i.e., the latent variables). This approach iteratively finds the expected values of the latent variables, and solves for the maximisation problem
\[
    \underset{\bm{\beta}}{\rm{arg max}}\; \E{}{\log p(\bm{\beta},\tau,\bm{\lambda}|{\bf y}) \vbar \bm{\beta}}
\]
to produce a sequence of estimates for the regression coefficients. This approach is effective because the scale-mixture form of the local-global prior means that conditional on $\bm{\lambda}$, the posterior for $\bm{\beta}$ is Gaussian and maximisation is (relatively) straightforward. In the case that the likelihood is non-Gaussian, but itself admits a scale-mixture-of-normals representation, such as the Laplace or logistic regression model, this approach can be adapted appropriately.\\

This approach is not easily applicable if the prior placed on $\bm{\beta}$ lacks a simple analytic form; in this case, the expected value of the shrinkage parameter will not have a closed form, and may be difficult or impossible to compute. One such prior is the horseshoe discussed previously. To attempt to address this problem~\cite{bhadra2019horseshoe}  introduced a ``horseshoe-like'' prior that mimics the behavior of the horseshoe density proposed by~\cite{carvalho2010horseshoe}. This horseshoe-like prior has a closed form density function and a scale-mixture representation that allows for implementation of an EM algorithm to obtain the ``horseshoe-like'' MAP estimates. 

\section{Bayesian EM Sparse Linear Regression} \label{sec: Bayesian EM sparse regression}
In this section, we present a novel, general EM procedure to compute the MAP estimate for the linear model. The key innovation underlying our approach is to treat the coefficients, $\bm{\beta}$, as latent variables; this is in contrast to the usual application of the EM algorithm~\cite{figueiredo2003adaptive} that treats the shrinkage hyperparameters, $\bm{\lambda}$, as missing data. \\

As is standard in penalised regression, and without any loss of generality, we assume that the predictors are standardised to have mean zero and standard deviation one, and the target has a mean of zero, i.e., the estimate of the intercept is simply $\hat{\beta}_0 = (1/n)\sum y_i$. This means we can simply ignore the intercept when estimating the remaining coefficients $\bm{\beta}$.

\subsection{The Basic EM Algorithm} \label{sec: Basic EM algorithm}
Here we consider the linear regression model define in hierarchy (\ref{eq:likelihood_regression})-(\ref{eq:global-local}).  The proposed EM algorithm solves for the posterior mode estimates by iterating through the following two steps until convergence is achieved:\\

\noindent \textbf{E-step}. We take the expectation of the complete negative log-posterior (with respect to the missing variable $\bm{\beta}$), conditional on the current values of $\bm{\lambda}$, $\tau^2$ and $\sigma^2$, and the observed data, ${\bf y}$; the resulting quantity is called the ``Q-function'':
\begin{align} 
&Q(\bm{\lambda}, \tau, \sigma^2|\hat{\bm{\lambda}}^{(t)}, \hat{\tau}^{(t)}, \hat{\sigma}^{2^{(t)}}) \nonumber \\
 &\; \; = \E{}{- \log p(\bm{\beta}, \bm{\lambda}, \tau, \sigma^2 \vbar {\bf y}) \: | \: \hat{\bm{\lambda}}^{(t)}, \hat{\tau}^{(t)}, \hat{\sigma}^{2^{(t)}}, {\bf y}} \nonumber \\
&\; \; = \left( \frac{n+p}{2} \right) \log \sigma^2 + \frac{\E{}{ ||{\bf y} - {\bf X}\bm{\beta}||^2 \:  | \: \hat{\bm{\lambda}}^{(t)}, \hat{\tau}^{(t)}, \hat{\sigma}^{2^{(t)}} }}{2 \sigma^2} + \frac{p}{2} \log \tau^2 \nonumber \\
& \; \; \; \; \; \; \; \; +\frac{1}{2}\sum_{j=1}^p \log \lambda_j^2 + \frac{1}{2 \sigma^2 \tau^2} \sum_{j=1}^p \frac{\E{}{\beta_j^2 \:  | \: \hat{\bm{\lambda}}^{(t)}, \hat{\tau}^{(t)}, \hat{\sigma}^{2^{(t)}} }}{\lambda_i^2} - \log \pi(\bm{\lambda}, \tau) \label{eq: E-step}
\end{align}
where $\pi(\bm{\lambda}, \tau)$ is the joint prior distribution for the hyperparameters. For notational simplicity we use $\E{}{\beta_j^2}$ and $\E{}{ || {\bf y} - {\bf X} \bm{\beta} ||^2 }$ for the conditional expectations of $\beta_j^2$  and the sum of squared residuals, respectively, throughout the sequel.\\

\noindent \textbf{M-step}. Update the parameter estimates by minimising the Q-function with respect to the shrinkage hyperparameters and noise variance, i.e.,
\begin{align} 
    \{ \hat{\bm{\lambda}}^{(t+1)}, \hat{\tau}^{(t+1)}, \hat{\sigma}^{2^{(t+1)}} \} = \argmin{\bm{\lambda},\tau,\sigma^2} \left\{ Q \left(\bm{\lambda}, \tau, \sigma^2 \vbar \hat{\bm{\lambda}}^{(t)}, \hat{\tau}^{(t)}, \hat{\sigma}^{2^{(t)}} \right) \right\}\label{eq: M-step}
\end{align}
Implementation of this EM algorithm requires only knowledge of the negative log-prior of choice $- \log \pi(\bm{\lambda}, \tau)$, the conditional expectations $\E{}{\bm{\beta^2}}$ and $\E{}{|| {\bf y}-{\bf X}\bm{\beta} ||^2}$. In Section \ref{sec: Estimating the conditional expectations}, we discuss several different approaches to compute the conditional expectations for the E-step. This algorithm is quite general, with only the M-step depending on the choice of prior for the coefficients $\beta_j$. Application to the specific case of the horseshoe is discussed in Section \ref{sec: Application to Horseshoe Estimator}. \\

\noindent The overall algorithm iterates the E-step and M-step until a convergence criterion is satisfied. Once convergence is achieved, we use the mode of the posterior distribution of $\bm{\beta}$, conditional on the final values of $\hat{\bm{\lambda}}^{(t)}$, $\hat{\tau}^{(t)}$ and $\hat{\sigma}^{2^{(t)}}$ as our point estimate. Given that this conditional distribution is Gaussian, the mode is just the mean of the normal distribution (\ref{eq:cond_post_beta}) given by (\ref{eq:exact_posterior_exp}). Finally we set components of the final conditional posterior mode  that are very small (smaller in absolute value than a small fraction of the standard error) to exactly zero. This step is technically not required, as given sufficient iterations the posterior mode will converge to be exactly sparse, but substantially reduces run time. In this paper we used $|\beta_j^{(t)}| < (5 \sqrt{n})^{-1}$ as the threshold for all experiments.

\subsection{Computing the conditional expectations}\label{sec: Estimating the conditional expectations}
The conditional expected values $\E{}{\bm{\beta^2}}$ and $\E{}{|| y-{\bf X}\bm{\beta} ||^2}$ depend on the conditional posterior distribution of the regression coefficients $\bm{\beta} \in \mathbb{R}^p$~\cite{makalic2016simple}:
\begin{align}\label{eq:cond_post_beta}
\begin{split}
    \bm{\beta}\,|\,\bm{\lambda},\tau,\sigma,{\bf y} \; &\sim \; N_p({\bf A}^{-1}{\bf X}^T{\bf y}, \sigma^2{\bf A}^{-1})\\
    {\bf A} \; &= \; ({\bf X}^T {\bf X} +  \tau^{-2} \bm{\Lambda}^{-1})
\end{split}
\end{align}
where ${\bm \Lambda} = \rm{diag}(\lambda^2_1, \cdots, \lambda^2_p)$. One important point to note here is that the posterior distribution of $\bm{\beta}$ does not depend on the choice of prior applied to $\lambda_j$. This implies that regardless of the (marginal) prior assigned to $\bm{\beta}$, as long as it has a SMN representation, both $p({\bf y}|\bm{\beta})$ and $\pi(\bm{\beta|\lambda},\tau)$ will always be Gaussian densities, e.g., it does not matter whether the prior assigned to $\bm{\beta}$ is a Laplace prior (to solve for the Lasso) or a horseshoe prior, the E-step will remain the same. Changing the marginal priors on the regression coefficients only requires straightforward modification to the M-step for updates on the shrinkage parameters (see Section \ref{sec: Application to Horseshoe Estimator}). This is an interesting advantage of our EM approach as the E-step is frequently very difficult to implement, particularly when dealing with conditional distributions that lack a standard density. 

\subsubsection{Exact expectations.}
The expected value of ${\beta}_j^2$ can be solved for by taking the sum of the variance and the square of the expected value of ${\beta}_j$:
\begin{align}
\begin{split}
    \E{}{{\beta}_j^2 \, | \, \bm{\lambda}^{(t)},  \tau^{(t)}} &= {\rm Var}[{\beta}_j] + \E{}{{\beta}_j}^2\label{eq:exact_Eb2}.
\end{split}
\end{align}
Due to the properties of Gaussian distributions, ${\rm Var}[{\beta_j}] = {\left( {\rm Cov}[\bm{\beta}] \right)}_{j,j}$, and
\begin{eqnarray}
    \nonumber
    {\rm Cov}[\bm{\beta}] &=& 
    \sigma^2 {
    \bf A}^{-1}, \\
    \label{eq:exact_posterior_exp}
    \E{}{\beta_j} &=& {\left( {\bf A}^{-1} {\bf X}^T {\bf y} \right)}_{j}.
\end{eqnarray}
Direct computation of this expectation value is potentially computationally expensive and numerically unstable because it involves inverting the $p \times p$ matrix, ${\bf A}$. The efficient algorithm  proposed in~\cite{rue2001fast} avoids explicitly computing the inverse of ${\bf A}$ when sampling from multivariate normal distributions of the form (\ref{eq:cond_post_beta}), and the approach in~\cite{bhattacharya2016fast} is similar but utilises the matrix inversion lemma for improved computational complexity when $p > n$. Both of these algorithms can be trivially modified to compute the exact mean and covariance matrix of the conditional posterior distribution of the hierarchy (\ref{eq:cond_post_beta}).\\

Similarly, finding $\E{}{|| {\bf y} - {\bf X} \bm{\beta} ||^2}$ involves solving for the expected value of a quadratic form of the regression coefficients, which in turns requires inverting the same matrix, ${\bf A}$. From standard results involving expectations of quadratic forms we have:
\begin{align}
\begin{split}
    \E{}{|| {\bf y}-{\bf X}\bm{\beta} ||^2} = || {\bf y} - {\bf X} \E{}{\bm{\beta}} ||^2 + {\rm tr}({\bf X}^T {\bf X} \cdot \rm{Cov}[\bm{\beta}]) \label{eq:exact_ERSS}
\end{split}
\end{align}
where ${\rm tr}(\cdot)$ is the usual trace operator.

\subsubsection{Approximate expectations.} \label{sec:Approximate expectation}
The two main components required to compute the expectations in the E-step for this linear model are the conditional posterior mean and variance of the regression coefficients $\bm{\beta}$ under the distribution (\ref{eq:cond_post_beta}). The conditional posterior mean can be found using the Rue's~\cite{rue2001fast} or Bhattacharya's~\cite{bhattacharya2016fast} algorithm without having to explicitly solve for the inverse of ${\bf A}$, with a simple extension of the same procedures allowing us to solve for the exact covariance matrix $\sigma^2 {\bf A} ^ {-1}$. However, the quantities that we need are the conditional variances (i.e., the diagonal elements of the conditional covariance matrix). Therefore, instead of inverting the full covariance matrix $\bf A$, the conditional variances may be approximately found by inverting only the diagonal elements of {\bf A}, i.e.,
\begin{align}
\begin{split}
    {\rm Var}[\beta_j] \; &\approx \; \sigma^2\,\frac{1}{A_{j,j}} \\
    &= \; \sigma^2\left( ||{\bf x}_j||^2 + \frac{1}{\tau^2 \lambda_j^2} \right)^{-1}. \label{eq:approx_var_lin}
\end{split}
\end{align}
where ${\bf x}_j$ is the $j$-th column of ${\bf X}$. This approximation takes only $O(p)$ operations. In the specific case that the predictors ${\bf X} = ({\bf x}_1, \cdots, {\bf x}_p)$ are orthogonal, ${\bf X}^T {\bf X} = c\, {\bf I}_p$ and the approximation (\ref{eq:approx_var_lin}) will recover the exact variance of $\beta_j$. This approximation also applies to (\ref{eq:exact_ERSS}), i.e., instead of computing the matrix multiplication of the components in the trace function, one may instead consider only the diagonal elements of the matrix, yielding the approximation
\begin{align}
\begin{split}
    {\rm tr}({\bf X}^T {\bf X} \cdot \rm{Var}[\bm{\beta}]) &\approx \; \sigma^2 \sum_{j = 1}^p \left( ||{\bf x}_j||^2 + \frac{1}{\tau^2 \lambda_j^2} \right)^{-1} ||{\bf x}_j||^2. \label{eq: approx_tr_lin}
\end{split}
\end{align}
In comparison to the $O(p^2)$ operations required to compute the trace of the product of matrices in (\ref{eq:exact_ERSS}), this approximation requires only $O(p)$ operations.

\section{Application to the Horseshoe Estimator} \label{sec: Application to Horseshoe Estimator}
\begin{figure}[h]
     \centering
     \begin{subfigure}[b]{0.3\textwidth}
         \centering
         \includegraphics[width=\textwidth]{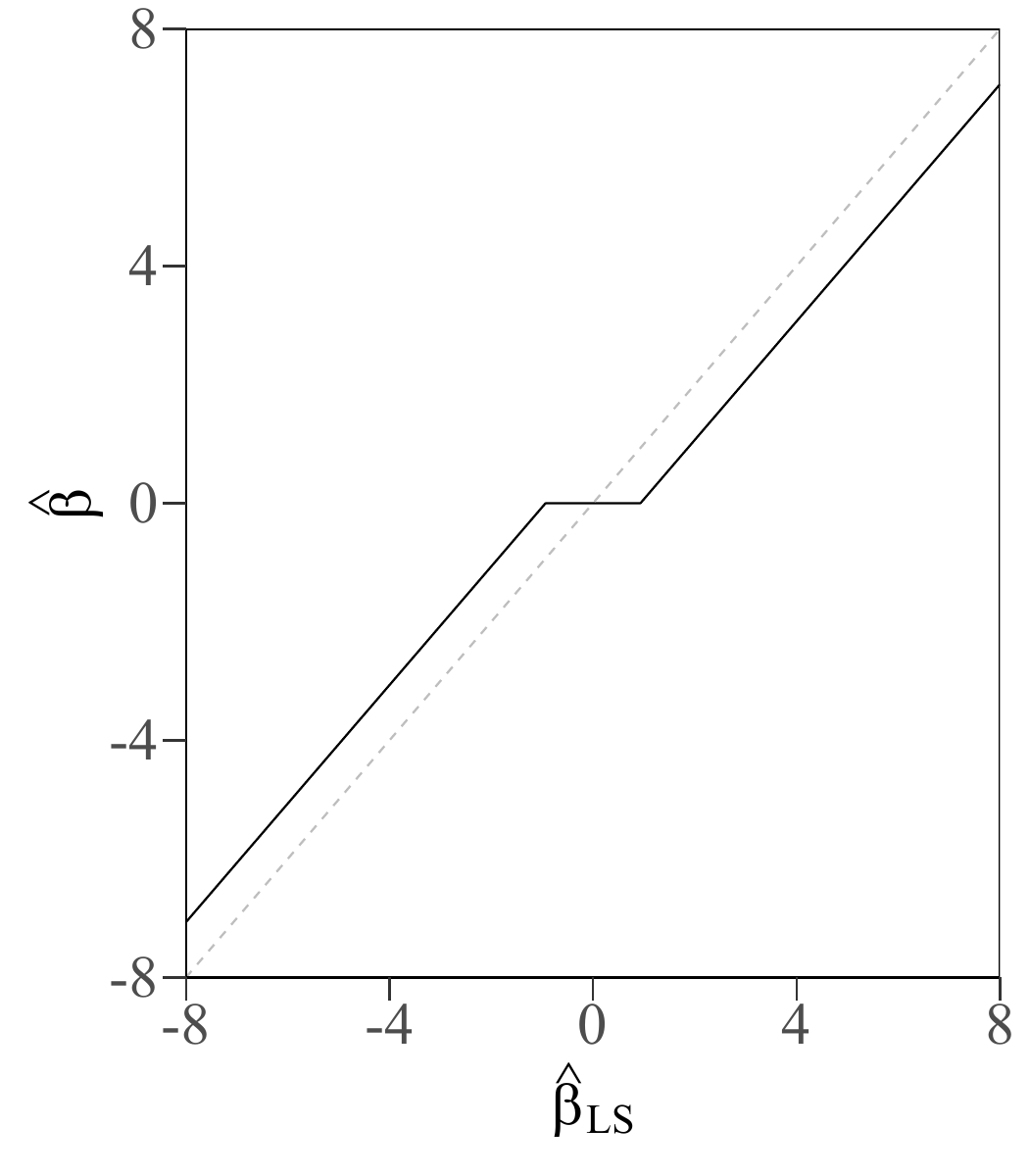}
         \caption{\footnotesize Lasso}
     \end{subfigure}
     \hspace{1em}%
     \begin{subfigure}[b]{0.3\textwidth}
         \centering
         \includegraphics[width=\textwidth]{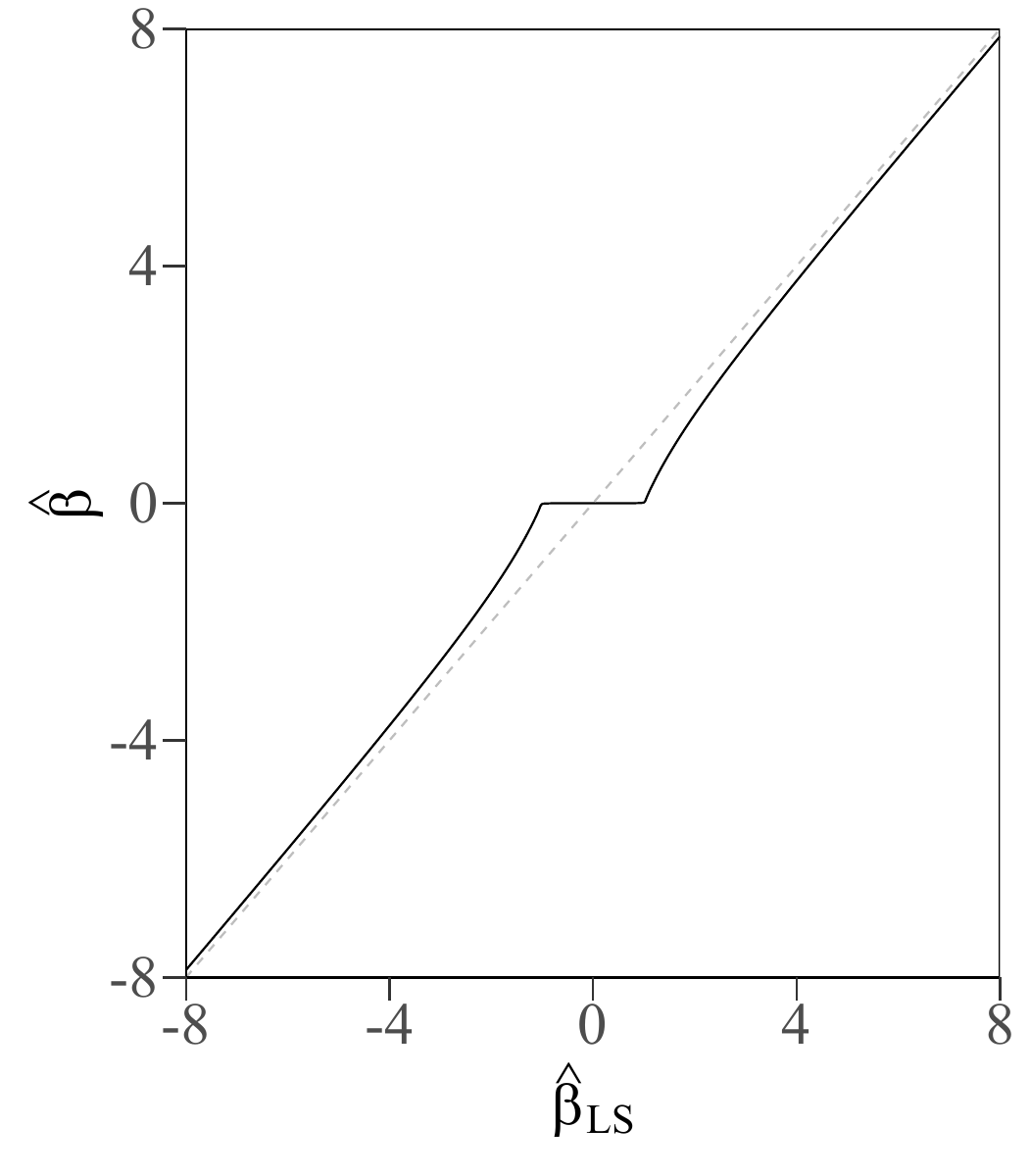}
         \caption{\footnotesize Horseshoe}
     \end{subfigure}
     \hspace{1em}%
     \begin{subfigure}[b]{0.3\textwidth}
         \centering
         \includegraphics[width=\textwidth]{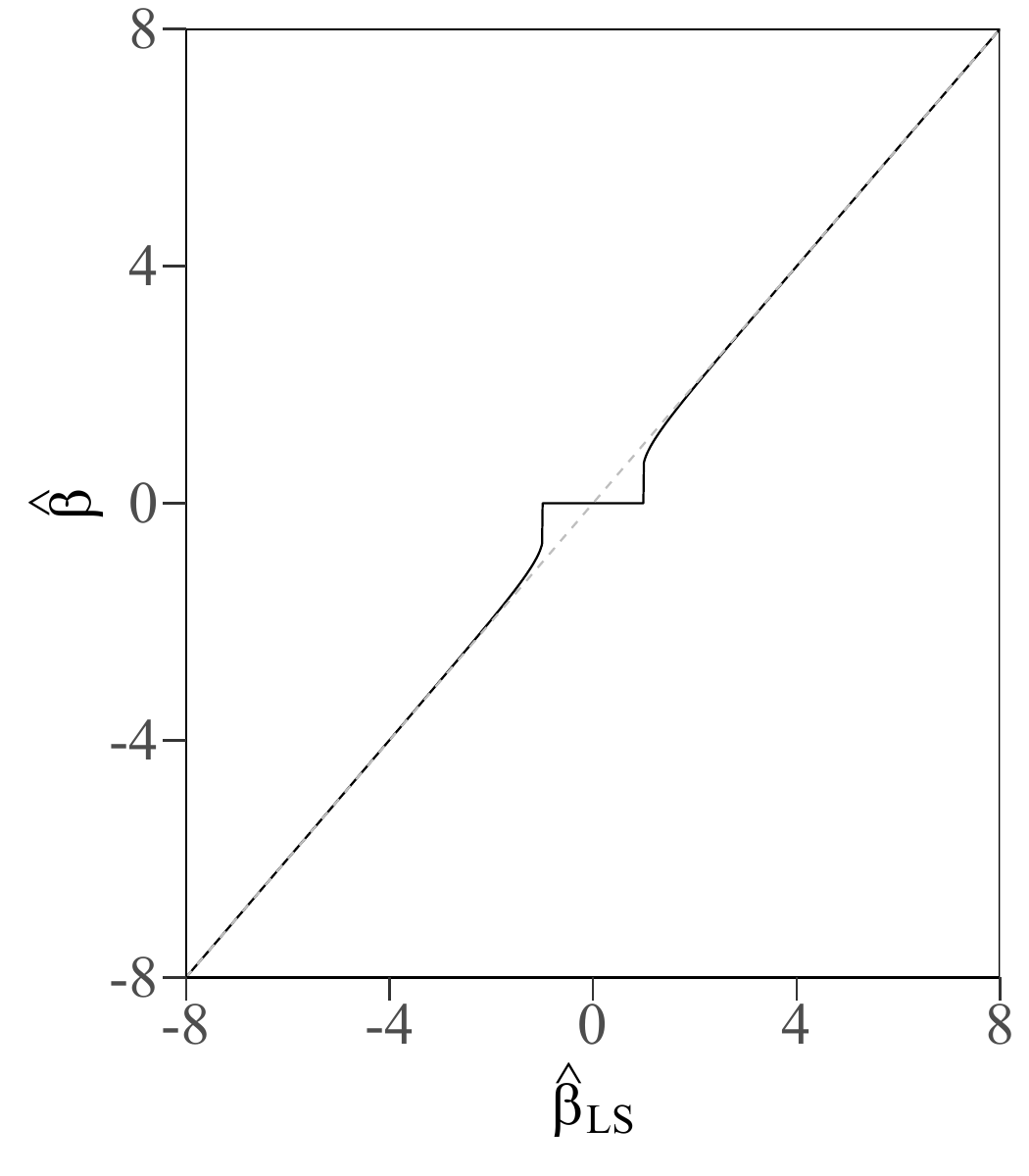}
         \caption{\footnotesize Horseshoe-like}
     \end{subfigure}
     \caption{\footnotesize The posterior mode estimates $\hat{\beta}$ versus $\hat{\beta}_{\rm LS}$ for the (a) Lasso, (b) Horseshoe and (c) Horseshoe-like estimator. For illustration purposes, $\tau$ is chosen such that all three estimators give nearly identical shrinkage within approximately 1 unit of the origin.}
     \label{fig: shrinkage_profile}
\end{figure}

We now demonstrate the application of the EM procedure described in Section \ref{sec: Basic EM algorithm} to the horseshoe prior and provide the exact EM updates for each of the shrinkage parameters. Here, we assume no prior knowledge on the sparsity of the regression coefficient and assign the recommended default prior for the global variance parameter~\cite{gelman2006prior,polson2012half}:
\begin{equation}
\tau \; \sim \; C^+(0,1), \; \; \tau \in (0,1)
\end{equation}
where $C^+ (0,1)$ denotes a standard half-Cauchy distribution with probability density function of
\begin{equation}
   p(z) = \frac{2}{\pi(1+z^2)}, \; z>0.
\end{equation}
We limit the range of $\tau$ to $(0,1)$ following~\cite{van2017adaptive}. The degree of sparsity applied to individual regression coefficients now depends on the choice of the prior distribution assigned to the local variance (shrinkage) component. Carvalho et al.~\cite{carvalho2010horseshoe} introduced the use of the horseshoe prior in sparse regression and demonstrated its robustness at handling sparsity with large signals. This is achieved by placing a half-Cauchy prior distribution over both the local and global shrinkage parameter. It was subsequently suggested that the horseshoe prior can be generalised by assigning an inverted-beta prior on $\lambda^2_j\;\sim\;\beta^{'}(a,b)$ with probability density function~\cite{polson2012half}
\begin{equation}\label{hier:GeneralisedHS}
    p(\lambda^2_j) = \frac{(\lambda^2_j)^{a-1} (1+\lambda^2_j)^{-a-b}}{B(a,b)}
\end{equation}
where $B(a,b)$ is the beta function, and $a>0$, $b>0$. We refer readers to~\cite{schmidt2019bayesian} for further details about the effect of these hyperparameter $a$ and $b$ on the origin and tail properties of the generalised horseshoe prior. Our interest is in the particular case that $a=b=1/2$ as this corresponds to a usual half Cauchy prior for $\lambda_j$. Substituting the negative logarithm of this prior distribution into (\ref{eq: E-step}) -- (\ref{eq: M-step}) and holding $\tau^2$ and $\sigma^2$ fixed yields the M-step update for each $\lambda_j^2$:
\begin{align}\nonumber
    \hat{\lambda}^2_j &= \underset{\lambda^2_j}{\rm{arg min}}\: \left\{\log \lambda_j^2 + \frac{W_j}{\lambda_j^2} + \log(1+\lambda_j^2) \right\} \\ 
     &=  \frac{1}{4} \left( \sqrt{1+6W_j+W_j^2}+W_j-1  \right) \label{eq: lambda_hs} \\
     (\hat{\tau}^2, \hat{\sigma}^2) &= \underset{\tau^2, \sigma^2}{\rm{arg min}}\: Q(\tau^2, \sigma^2; \hat{\bm{\lambda}^2}) \label{eq: optim_tausig}
\end{align}
where $W_j ={E[\beta_j^2]}/{(2\sigma^2\tau^2)}$. For convenience, the negative log-prior and the M-step update for the horseshoe prior are summarised in Table \ref{tab:EMupdates}. Given the $\hat{\bm{\lambda}}^2$ updates (\ref{eq: lambda_hs}), the $\hat{\tau}$ and $\hat{\sigma}^2$ estimates can be found using numerical optimisation. This two-dimensional optimisation problem~(\ref{eq: optim_tausig}) can be further reduced to the following one-dimensional optimisation problem by (approximately) estimating $\sigma^2$ using the expected residual sum-of-squares given in Equation (\ref{eq:exact_ERSS}) or (\ref{eq:approx_var_lin}):
\begin{align}\nonumber
    \hat{\sigma}^2 &= \E{}{|| {\bf y}-{\bf X}\bm{\beta} ||^2}/n, \\
     \hat{\tau}^2 &= \underset{\tau^2}{\rm{arg min}}\: Q(\tau^2; \hat{\bm{\lambda}}^2,\hat{\sigma}^2).
\end{align}
This approximate update for $\sigma^2$ is within $O(n^{-1})$ of the exact solution, and allows us to substantially reduce the complexity of the optimisation problem. Figure \ref{fig: shrinkage_profile} compares the shrinkage profiles for the Lasso, horseshoe and horseshoe-like estimator. The plots demonstrate how the three estimators shrink the least-square (unpenalised) estimate $\hat{\beta}_{\rm LS}$ towards zero. All three procedures shrink small least-squares estimates to zero. The lasso affects all values of $\hat{\beta}_{\rm LS}$ by translating them towards zero by the same amount, while the horseshoe and horseshoe-like leave large values of $\hat{\beta}_{\rm LS}$ largely unshrunk. The horseshoe-like prior exerts more of a hard thresholding-like behaviour as $\hat{\beta}_{\rm LS}$ approaches zero, while the horseshoe prior mimics firm thresholding.
\begin{table*}[t]
\vspace{-4mm}
\caption{\footnotesize Negative log-prior distribution and EM updates for the proposed horseshoe estimator, where $W_j ={E[\beta_j^2]}/{(2\sigma^2\tau^2)}$.}
\label{tab:EMupdates}
\begin{center}
\begin{tabular}{@{\hskip 0.1in}c@{\hskip 0.7in}c@{\hskip 0.1in}}
\toprule
$- \log \pi(\bm{\lambda}, \tau)$ & M-step\\
\midrule
$\displaystyle \sum_{j=1}^p \left[ \log(1+\lambda_j^2) + \frac{\log \lambda_j^2}{2} \right] + \log(1+\tau^2)$ & $\left.\hat{\lambda}^2_j\right.^{(t+1)} =  \displaystyle{ \frac{{\left(\sqrt{1+6W_j+W_j^2}\right)+W_j-1} }{{4}} }$\\
\bottomrule
\end{tabular}   
\end{center}
\vspace{-4mm}
\end{table*}

\section{Extension to Generalised Linear Models} \label{sec: Extension to GLM}

We now consider an extension of the EM approach proposed in Section~\ref{sec: Bayesian EM sparse regression} to non-normal data models through the framework of generalised linear models (GLMs)~\cite{nelder1972generalized}. A GLM models the conditional mean of the target by an appropriate (potentially) non-linear transformation of the linear predictor; well known examples include binomial logistic regression and Poisson log-linear regression. In general, when the targets are non-Gaussian, the conditional distribution of the coefficients will not have a standard form, and finding the exact expectations $\E{}{\beta_j^2}$ is difficult. However, it is frequently the case that the conditional distribution can be approximated by a heteroskedastic Gaussian distribution
\begin{align}\label{eq:cond_post_beta:glm}
\begin{split}
    \bm{\beta} \vbar \bm{\lambda}, \tau, \bm{\omega}, {\bf z} \; &\sim \; N_p\left( {\bf A}_{\bm{\omega}}^{-1} {\bf X}^T \bm{\Omega} {\bf z}, \; {\bf A}_{\bm{\omega}}^{-1} \right),\\
    {\bf A}_{\bm{\omega}} \; &= \; {\bf X}^T \bm{\Omega} {\bf X} + \tau^{-2} \bm{\Lambda}^{-1}
\end{split}
\end{align}
where $\bm{\Omega} = {\rm diag}(\bm{\omega})$, $\bm{\omega} = (\omega_1,\ldots,\omega_n)$ is a vector of weights, and ${\bf z}$ is an adjusted version of the targets, ${\bf y}$. Via standard central-limit theorem arguments, the accuracy of this approximation increases as the sample size $n$ grows. The weights can be obtained via a linearisation argument (i.e., the well known IRLS algorithm) or, preferably, via a scale-mixture of normals representation of the likelihood when available. For example, the logistic regression implementation used in this paper utilises the well-known Poly\'{a}-gamma representation of logistic regression~\cite{polson2013bayesian}; under this scheme, the adjusted targets and weights are
\[
    z_i = (y - 1/2) / \omega_i, \; \; \omega_i = \left( \frac{1}{2 \eta_i} \right) \tanh \left( \frac{\eta_i}{2} \right),
\]
where $\bm{\eta} = {\bf X} \bm{\beta} + \beta_0 {\bf 1}_n$ is the linear predictor.
Given appropriate weights and adjusted targets, one may simply approximate the conditional posterior-covariance of the coefficients by $ {\rm Cov}\left[ \bm{\beta} \right] \approx {\bf A}^{-1}_{\bm{\omega}}$ and use either (\ref{eq:exact_posterior_exp}) or (\ref{eq:approx_var_lin}) to obtain approximate expressions for $\E{}{\beta_j^2}$. Alternatively, one could potentially utilise a stochastic variant of the EM algorithm~\cite{celeux1985sem} to compute the required expectations, but due to space constraints we do not consider this further in this paper.

\section{Experimental Results} \label{sec: results}

We compare the performance of our proposed EM-based horseshoe estimator (HS-EM, using the exact expectations, and HS-apx, using the approximate expectations) against several state-of-the-art sparse methods including non-convex estimators (SCAD, MCP, HS-like estimator~\cite{bhadra2019horseshoe}), lasso and ridge estimator. We analyse the performance of the estimators in terms of variables selected and prediction accuracy on both simulated and real data. The experiments in this section are designed for variable selection under high dimensional settings, and we use the following metrics:
\begin{itemize}
\footnotesize
    \item \textbf{MSE}. Mean squared prediction error; given by  $(\hat{\bm{\beta}} - \bm{\beta})'\bm{\Sigma}(\hat{\bm{\beta}} - \bm{\beta})$ for simulated data experiments, and $(1/n)||\hat{{\bf y}} - {\bf y} ||^2_2$ for real data experiments.
    \item \textbf{Time}. Computation time (in seconds).
    \item \textbf{No.V}. Number of variables included in the model.
    \item \textbf{TNZ}. True non-zeros. Number of non-zero coefficients correctly identified.
    \item \textbf{FNZ}. False non-zeros. Number of zero coefficients incorrectly identified as non-zero.
\end{itemize}
All experiments are performed in the R statistical platform. Datasets and code for the experimental results in this section are publicly available\footnote{Available at https://github.com/shuyu-tew/Sparse-Horseshoe-EM.git}. The proposed EM horseshoe estimator is also implemented in the \textbf{bayesreg}\footnote{Available at \url{https://cran.r-project.org/web/packages/bayesreg/index.html}} Bayesian regression package. We use the \texttt{glmnet} package to implement the lasso and ridge, while the \texttt{ncvreg} package implements both MCP and SCAD. The hyperparameters of the non-Bayesian techniques are tuned using ten-fold cross-validation. All function arguments are set to the default value unless mentioned otherwise. Our proposed estimator terminates when it satisfies the convergence criterion: $\sum_{j=1}^p(|\beta_j^{(t)} - \beta_j^{(t+1)}|)/ (1+ \sum_{j=1}^p(|\beta_j^{(t+1)}|)) < \omega$, where $\beta^{(t)}$ denotes the coefficient estimates at iteration $t$ and $\omega$ is the tolerance parameter which we set it to $10^{-5}$. 

\subsection{Simulated Data} \label{sec: Simulated Data}
This section compares the prediction and variable selection performance of various sparse estimators with the normal means model and linear regression model across various simulation settings. The experimental setups are based on the simulation study presented in \cite{bhadra2019horseshoe}.

\subsubsection{Normal Means Model} \label{sec: Normal Means}
Here, $n = 1000$ with the true sparse mean vector, $\bm{\beta} = \{\bm{b}_{10},\bm{-b}_{10},\bm{0}_{980}\}$ where $b = \{3, 10\}$, $\bm{b}_{10}$ is a vector of length 10 with entries equal to $b$, while $\bm{0}_{980}$ is a vector of length 980 with entries equal to $0$. The data is generated as $(y_i | \beta_i) \; \sim \; N(\beta_i,1) $, for which $\bm{y}$ is generated from a normal distribution with mean $\bm{\beta}$ and a variance of 1. We repeat this simulation for 100 times and the results are summarised in Table \ref{tab:mm_results}. It should be noted that the implementations described in Section \ref{sec: Bayesian EM sparse regression} is built upon a regression model. This normal means model can be seen as a special case of the regression model with $p = n$ and $X = I_{n \times n}$ where $I_{n \times n}$ is an identity matrix of order n. But naively using the methods described in Section \ref{sec: Bayesian EM sparse regression} to solve for this normal means problem is not ideal and slow, because we will have to work with matrix of order 1000. There will be no changes to the M-step of our EM algorithm since there is no change to the hierarchy of the priors. The only difference in implementation is the computation of the conditional expected distribution (see Appendix~\ref{apx: conditional expectation}).\\

Table \ref{tab:mm_results} compares the performance for our proposed horseshoe estimator against the HS-Like estimator, the VisuShrink~\cite{donoho1994ideal} estimates and the BayesShrink~\cite{chang2000adaptive} estimates of the lasso. Overall, the HS posterior mode estimates is competitive with or superior to the lasso estimates and outperforms the horseshoe-like estimator in terms of MSE and the number of correctly identified zero coefficient. Most of the variables included in the HS-like estimator are zero coefficients incorrectly identified as non-zeros. This suggests model overfitting.

\begin{table*}[b]
\caption{Performance of different sparse estimators (with associated standard error in parentheses) for the simulated data experiments on normal means model. $\rm{Lasso_{BS}}$ and $\rm{Lasso_{VS}}$ are the VisuShrink and BayesShrink estimates of the lasso respectively.}
\small
\centering
    \begin{tabular}{l@{\hskip 0.1in}c@{\hskip 0.05in}c@{\hskip 0.05in}c@{\hskip 0.05in}c@{\hskip 0.05in}c@{\hskip 0.2in}c@{\hskip 0.05in}c@{\hskip 0.05in}c@{\hskip 0.05in}c}
    \toprule[1pt]
       & HS-EM & HS-like & $\rm{Lasso_{BS}}$ & $\rm{Lasso_{VS}}$ & & HS-EM & HS-like & $\rm{Lasso_{BS}}$ & $\rm{Lasso_{VS}}$ \\
    \midrule
    \\[-0.5em]
    & \multicolumn{4}{c}{$\bm{\beta} = \{{\bf 3}_{10},{\bf -3}_{10},{\bf 0}_{980}\}$} & & \multicolumn{4}{c}{$\bm{\beta} = \{{\bf 10}_{10},{\bf -10}_{10},{\bf 0}_{980}\}$}\\
    \\[-0.5em]
MSE         & 148.6(1.6) & 882.3(9.06) & {\bf 115.3}(2.3) & 165.9(0.7) && {\bf 26.41}(1.6) & 883.3(9.04) & 309.1(2.9) & 296.5(3.3)\\
Time        & 0.38(0.01) & 0.003(0.00)  & 0.00(0.00) & 0.00(0.00) && 0.15(0.01) & 0.002(0.00) & 0.00(0.00) & 0.00(0.00)\\
$\rm{Inc_z}$& 0.07(0.03) & 528.4(25.2) & 25.44(2.1) & 0.17(0.05) && 0.07(0.03) & 534.6(25.6) & 471.6(2.3) & 0.17(0.05)\\
No.V        & 3.86(0.19) & 548.2(25.2) & 39.5(2.39) & 4.99(0.19) && 20.07(0.5) & 554.6(25.6) & 491.6(2.3) & 20.17(0.1)\\
    \\[-0.5em]
    \bottomrule[1pt]
    \end{tabular}
    \label{tab:mm_results}
\end{table*}

\subsubsection{Linear Regression Model}\label{sec: Linear model}
We considered $n = 70$, $p = 350$ and simulated 100 different data sets from the linear model, $ {\bf y} \; \sim \; N_n\left({\bf X}\bm{\beta}, \; \sigma^2\bm{I}_n\right)$ with $\sigma^2 \in \{1, 9\}$ . The predictor matrix, $\bm{X} \in \mathbb{R}^{n \times p}$ was generated from a correlated multivariate normal distribution $N_p(0, \bm{\Sigma})$ with $\Sigma_{ij} = \rho^{|i-j|}$ and $\rho \in \{0, 0.7\}$. \\

The first ten entries of $\bm{\beta}$ were set to $3$, the next ten were set to $-3$, and the remaining entries were zero. The results are shown in Table \ref{tab:simulated_results}. Overall, the HS-like estimator performed the best, obtaining the lowest MSE at the expense of substantial overfitting. The good performance in MSE under this setting is not unexpected, as  this problem setting favors estimators that overfit (i.e., are less conservative) to a certain extent. Given a sparse vector with only 20 non-zero coefficients, this setting in which all non-zero coefficients have the same magnitude is one of the hardest settings for a sparse method, as excluding any one of the true coefficients is equally damaging. The prediction risk of this problem is $\rm{E}[|| \bm{\beta}-\hat{\bm{\beta}}||^2]$; if a method incorrectly includes an irrelevant predictor in the model, the risk will approximately be the variance of the estimate associated with this variable, i.e., ${\rm E}[\hat{\beta}_j^2] \approx \sigma^2/n$; in contrast, incorrectly excluding non-zero predictors will incur an error of $3^2$, which is much larger than $\sigma^2/70$ even for the setting of greater noise, i.e., $\sigma^2=9$. In general, approximately half of the coefficients included in the HS-like model are false non-zeros (FNZ); while most of the coefficients included in our HS models are true non-zeros with consistent small difference between TNZ and No.V.\\
\begin{table*}[t]
\caption{ Performance of different sparse estimators (with associated standard error in parentheses) for the simulated data experiments.}
\scriptsize
\centering
    \begin{tabular}{@{}l@{\hskip 0.1in}c@{\hskip 0.08in}c@{\hskip 0.08in}c@{\hskip 0.08in}c@{\hskip 0.08in}c@{\hskip 0.08in}c@{\hskip 0.08in}c@{\hskip 0.08in}c@{}}
    \toprule[1pt]
       & HS-EM & HS-apx & HS-like & Lasso & MCP & SCAD & Ridge\\
    \midrule
    \\[-0.8em]
    &\multicolumn{7}{c}{($\rho = 0, \sigma^2 = 1$)}\\
    \\[-0.8em]
MSE     & 162.7(3.5) & 162.8(2.9) & {\bf 119.8}(4.7) & 139.1(3.4) & 159.6(1.9) & 149.8(1.8)& 176.6(0.5)\\
Time    & 5.93(0.19) & 1.37(0.04) & 0.27(0.02) & 0.14(0.01) & 0.21(0.01) & 0.32(0.01) & 0.64(0.01) \\
No.V    & 6.11(0.35) & 5.83(0.36) & 43.4(0.53) & 21.4(1.59) & 8.61(0.52) & 18.7(0.72) & 350(0.00)\\
TNZ     & 4.37(0.29) & 4.27(0.28) & 14.2(0.32) & 8.77(0.61) & 4.89(0.29) & 8.31(0.32) & 20.0(0.00)\\
FNZ     & 1.63(0.16) & 1.56(0.16) & 29.2(0.72) & 12.6(1.06) & 3.72(0.28) & 10.4(0.51) & 330(0.00) \\
    \cmidrule{2-8}
    \\[-0.8em]
    &\multicolumn{7}{c}{($\rho = 0, \sigma^2 = 9$)}\\
    \\[-0.8em]
MSE     & 171.6(3.1) & 170.2(2.9) & {\bf 141.5}(3.5) & 151.9(2.7) & 164.2(1.9) & 153.6(1.9) & 177.1(0.5)\\
Time    & 5.89(0.22) & 1.37(0.04) & 0.27(0.01) & 0.15(0.01) & 0.21(0.01) & 0.33(0.01) & 0.64(0.01) \\
No.V    & 5.38(0.34) & 5.19(0.34) & 46.3(0.29) & 15.9(1.47) & 7.83(0.49) & 18.4(0.77) & 350(0.00)\\
TNZ     & 3.69(0.27) & 3.68(0.27) & 13.2(0.31) & 6.61(0.54) & 4.31(0.28) & 7.58(0.31) & 20.0(0.00)\\
FNZ     & 1.69(0.17) & 1.51(0.15) & 33.1(0.36) & 9.35(0.99) & 3.52(0.29) & 10.8(0.56) & 330(0.00)\\
    \cmidrule{2-8}
    \\[-0.8em]
    &\multicolumn{7}{c}{($\rho = 0.7, \sigma^2 = 1$)}\\
    \\[-0.8em]
MSE     & 12.7(1.29) & 15.1(1.57) & {\bf 3.79}(0.57) & 6.48(0.47) & 78.1(3.23) & 74.3(3.01) & 426.6(3.81) \\
Time    & 2.83(0.21) & 0.97(0.05) & 0.22(0.01) & 0.11(0.01) & 0.14(0.01) & 0.16(0.01) & 0.65(0.01)\\
No.V    & 16.9(0.25) & 16.4(0.29) & 25.1(0.46) & 33.4(0.49) & 14.1(0.33) & 22.4(0
49) & 350(0.00) \\
TNZ     & 16.9(0.25) & 16.4(0.29) & 19.7(0.05) & 19.7(0.06) & 8.21(0.16) & 9.64(0.16) & 20.0(0.00)\\
FNZ     & 0.04(0.03) & 0.04(0.03) & 5.27(0.51) & 13.7(0.48) & 5.89(0.31) & 12.8(0.49) & 330(0.00) \\
    \cmidrule{2-8}
    \\[-0.8em]
    &\multicolumn{7}{c}{($\rho = 0.7, \sigma^2 = 9$)}\\
    \\[-0.8em]
MSE     & 34.8(1.79) & 38.1(1.91) & 36.3(1.38) & {\bf 27.1}(1.17) & 85.9(3.28) & 84.8(3.21) & 428.7(3.89) \\
Time    & 4.41(0.26) & 1.17(0.04) & 0.24(0.01) & 0.12(0.01) & 0.14(0.01) & 0.16(0.01) & 0.64(0.01) \\
No.V    & 12.9(0.27) & 12.3(0.28) & 40.9(0.28) & 28.6(0.62) & 13.3(0.31) & 21.1(0.45) & 350(0.00)\\
TNZ     & 12.6(0.27) & 12.1(0.27) & 18.7(0.11) & 18.4(0.08) & 7.66(0.13) & 8.85(0.16) & 20.0(0.00)\\
FNZ     & 0.24(0.05) & 0.24(0.05) & 22.2(0.29) & 10.3(0.59) & 5.61(0.28) & 12.2(0.42) & 330(0.00)) \\
    \\[-0.8em]
    \bottomrule[1pt]
    \end{tabular}
    \label{tab:simulated_results}
    \vspace{-4mm}
\end{table*}

The HS-EM, SCAD and MCP methods all generally select much fewer variables than the HS-like estimator, with the HS-EM method overfitting less than HS-like and underfitting more than MCP or SCAD. In terms of MSE, our HS-EM method is competitive with, or superior to SCAD and MCP for all settings, despite producing substantially sparser estimates. It appears that both SCAD and MCP are sensitive to the correlation structure of the data; their performance is comparable to our HS-EM approach when there is no correlation, but when correlation is introduced to the data, both methods performed substantially worse in terms of MSE. Such behaviour is not observed in our proposed HS estimator. Of the two proposed HS estimators, HS-apx appears to be approximately three times faster than HS-EM. Otherwise, both estimators give roughly similar performance.
\begin{table*}[b]
\vspace{-4mm}
\caption{\footnotesize The posterior mode and mean estimates of the predictors for the diabetes data using the horseshoe estimator. The posterior mean and 95\% credible intervals are computed using the \texttt{bayesreg} package in R.}
\centering
    \begin{tabular}{l@{\hskip 0.12in}c@{\hskip 0.12in}c@{\hskip 0.12in}c@{\hskip 0.12in}c@{\hskip 0.12in}c@{\hskip 0.12in}c@{\hskip 0.12in}c@{\hskip 0.12in}c@{\hskip 0.12in}c@{\hskip 0.12in}c}
    \toprule[1pt]
    & AGE & SEX & BMI & BP & S1 & S2 & S3 & S4 & S5 & S6\\
    \midrule
    \\[-1.2em] 
    Mean    & -0.009 & -18.68 & 5.769 & 1.034 & -0.223 &  0.013 & -0.592 &  2.419 & 48.84 & 0.179 \\
    2.50\%  & -0.341 & -30.93 & 4.371 & 0.571 & -0.937 & -0.342 & -1.415 & -3.462 & 32.24 & -0.225\\
    97.5\%  &  0.326 & -5.144 & 7.109 & 1.457 &  0.098 &  0.656 &  0.189 &  11.36 & 70.14 & 0.734\\
    \\ [-0.5em]
    Mode    & $\cdot$ & -17.54 & 5.741 & 1.021 & $\cdot$ & $\cdot$ & -0.909 & $\cdot$ & 43.58 & $\cdot$ \\
    \\[-1.2em]
    \bottomrule[1pt]
    \end{tabular}
    \label{tab:diabetes_coef}
    \vspace{-4mm}
\end{table*}

\subsection{Real data}
We further analyse the performance of our proposed method on six real-world datasets. All datasets are available for download from the UCI machine learning repository~\cite{asuncion2007uci} unless mentioned otherwise. Each dataset is randomly split such that the training data has a sample size of $n$ with the remaining $N-n$ datapoints used as testing samples. This procedure is repeated 50 times and the averaged summary statistics for the performance measures are presented in Tables \ref{tab:real_results} and \ref{tab:real_results_logistic} for the linear regression and logistic regression experiments, respectively.

\subsubsection{Linear Regression}
Table~\ref{tab:diabetes_coef} presents the posterior mean estimates and their 95\% credible intervals (CI) for the diabetes data~\cite{efron2004least} ($N = 442, P = 10$). For comparison, we included the posterior mode estimates computed from the proposed HS-EM estimator. The HS-EM posterior mode estimates are very similar to the corresponding posterior means, and all posterior mode estimates are within the corresponding $95\%$ credible intervals; however, the HS-EM posterior mode provides a sparse point estimate as it excludes all variables (except S3) with 95\% CIs that include zero.\\

In addition to the diabetes data, we also analysed the benchmark Boston housing data ($N = 506, P = 14$), the concrete compressive strength dataset ($N = 1030, P = 9$) and the eye data ($N = 120, P = 200$)~\cite{scheetz2006regulation}. To make the problem more difficult, for each dataset we added a number of noise variables generated using the same procedure described in Section~\ref{sec: Simulated Data}; in all cases we added $p = 15$ additional noise variables, with $\rho = 0.8$. Model fitting is done using all the $P + 15$ (original plus noise) predictors as well as all interactions and possible transformations of the variables (logs, squares and cubics).\\

Overall, our proposed HS estimator performs the best, attaining the lowest MSE on all real data, while generally selecting the simplest models (included the lowest number of variables). Similar to the results observed in Section~\ref{sec: Simulated Data}, the results for HS-apx are virtually identical to the exact HS-EM procedure, while being substantially faster. SCAD and MCP exhibit comparable performance to HS in terms of prediction error for the diabetes data and eye data, but have relatively poor performance for Boston and concrete data. Interestingly, while HS-like dominated the other non-convex estimators in terms of MSE on the simulated experiments, it performs quite poorly on the real data analysis, particularly on the diabetes and Boston housing data. The HS-like estimator tended to include the highest number of variables, suggesting it is potentially prone to overfitting as discussed in Section~\ref{sec: Simulated Data}. Ridge regression generally performs worse than the sparse estimators, as would be expected given the experimental design.
\begin{table*}[t]
\vspace{-4mm}
\caption{\footnotesize Performance of different sparse estimators (with associated standard error in parentheses) on the 4 datasets for linear regression model.}
\small
\centering
    \begin{tabular}{@{}l@{\hskip 0.03in}c@{\hskip 0.04in}c@{\hskip 0.04in}c@{\hskip 0.04in}c@{\hskip 0.04in}c@{\hskip 0.04in}c@{\hskip 0.04in}c@{\hskip 0.04in}c@{}}
    \toprule[1pt]
       & HS-EM & HS-apx & HS-like & Lasso & MCP & SCAD & Ridge \\
    \midrule
    \\[-0.8em]
    \multicolumn{8}{c}{\textbf{Diabetes} ($n = 100, p = 385$)}\\
    \\[-0.8em]
    MSE     & {\bf 3383.3}(30.5)& 3407.2(32.1) & 13068(491.1) & 3654.4(47.9) & 3624.4(66.2) & 3667.6(68.5) & 4181.6(42.8)\\
    Time    & 3.01 (0.02) & 1.02 (0.02) & 0.32 (0.01) & 0.29 (0.01) & 1.13 (0.01) & 1.47 (0.02) & 0.74 (0.01)\\
    No.V    & 1.62 (0.09) & 1.54 (0.09) & 95.3 (0.29) & 4.14 (0.44) & 3.68 (0.41) & 6.86 (0.74) & 385 (0.00)\\
    \cmidrule{2-8}
    \\[-0.8em]
    \multicolumn{8}{c}{\textbf{Boston Housing} ($n = 100, p = 473$)}\\
    \\[-0.8em]
    MSE  & 26.76(0.71) & {\bf 26.72}(0.71) & 58.01(3.05) & 31.41(0.89) & 293.1(24.3) & 323.9(29.7) & 49.19(1.19) \\
    Time & 4.74(0.34) & 1.86(0.03) & 0.67(0.03) & 0.20(0.01) & 1.21(0.01) & 1.39(0.01) & 1.06(0.01) \\
    No.V & 2.82(0.16) & 2.84(0.16) & 47.9(0.78) & 4.60(0.52) & 4.52(0.41) & 10.9(0.87) & 473(0.00) \\
    \cmidrule{2-8}
    \\[-0.8em]
    \multicolumn{8}{c}{\textbf{Concrete} ($n = 100, p = 327$)}\\
    \\[-0.8em]
    MSE  & {\bf 73.71} (3.67) & 73.76 (3.66) & 222.4 (8.38) & 81.67 (1.98) & 113.9 (15.9) & 108.3 (16.1) & 176.2 (2.49)\\
    Time & 2.39 (0.17) & 0.86 (0.03) & 0.36 (0.02) & 0.23 (0.01) & 0.98 (0.01) & 1.14 (0.01) & 0.55 (0.01)\\
    No.V & 5.46 (0.13) & 5.42 (0.14) & 67.8 (0.57) & 10.7 (0.61) & 6.60 (0.44) & 13.8 (0.85) & 327 (0.00)\\
    \cmidrule{2-8}
    \\[-0.8em]
    \multicolumn{8}{c}{\textbf{Eye} ($n = 100, p = 200$)}\\
    \\[-0.8em]
    MSE  & \textbf{0.79} (0.06) & \textbf{0.79} (0.06) & 1.50 (0.23) & 0.84 (0.14) & \textbf{0.79} (0.06) & 0.86 (0.09) & 0.85 (0.09)\\
    Time & 0.52 (0.02) & 0.24 (0.01) & 0.15 (0.01) & 0.20 (0.01) & 0.13 (0.01) & 0.19 (0.01) & 0.23 (0.01)\\
    No.V & 3.84 (0.11) & 3.72 (0.09) & 0.00 (0.00) & 18.0 (0.55) & 5.20 (0.27) & 10.5 (0.42) & 200 (0.00)\\
    \\[-0.8em]
    \bottomrule[1pt]
    \end{tabular}
    \label{tab:real_results}
    \vspace{-4mm}
\end{table*}
\subsubsection{Logistic Regression}
We also tested the HS-EM estimator on two binary classification problems: the Pima indians data ($N = 768, P = 8$) and the heart disease data ($N = 302, P = 14$). Similar to the linear regression analysis, we augment the data with 10 noise variables and model the predictors together with their interactions and transformations. For the heart disease data, we only included the interactions between the variables and not the transformations, as there were a substantial number of categorical predictors. For this experiment we varied the number of training samples $n \in \{100, 200\}$.\\

The HS-EM estimator obtains the best classification accuracy and negative log-loss on the Pima data for both training sample sizes. Ridge regression, on the other hand, performs the best in terms of classification accuracy for the heart data. This suggest that many of the predictors and interactions between the variables in the heart data are potentially correlated with the response variable. Therefore, a sparse estimator might not be suitable for modelling this data. Nonetheless, our HS methods gives comparable results to the other sparse methods in terms of log-loss and in general, as the sample size increases, the difference in performance between all estimators becomes less significant.
\begin{table*}[t]
\caption{\footnotesize Performance of different sparse estimators on the Pima and heart disease data. CA is the classification accuracy and NLL is the negative log-loss.}
\small
\centering
    \begin{tabular}{@{}l@{\hskip 0.1in}l@{\hskip 0.1in}l@{\hskip 0.08in}c@{\hskip 0.08in}c@{\hskip 0.08in}c@{\hskip 0.08in}c@{\hskip 0.08in}c@{\hskip 0.08in}c@{\hskip 0.08in}c@{}}
    \toprule[1pt]
    Dataset & & HS-EM & HS-apx & Lasso & MCP & SCAD & Ridge\\
    \midrule
    \\[-0.8em]
    & & \multicolumn{6}{c}{($n = 100, p = 214$)}\\
    \\[-0.8em]
    \multirow{10}{*}{\textbf{Pima}}
    & CA        & {\bf 0.743}(0.01) & 0.741(0.01) & 0.704(0.01) & 0.741(0.01) & 0.735(0.01) & 0.674(0.01) \\
    & NLL   & {\bf 358.1}(2.42) & 358.8(2.55) & 381.5(3.33) & 360.9(4.12) & 362.8(3.90) & 400.7(2.63) \\
    & Time      & 0.622(0.02) & 0.533(0.01) & 0.298(0.01) & 1.475(0.07) & 2.056(0.07) & 0.251(0.01)\\
    & No.V      & 1.760(0.09) & 1.700(0.09) & 3.780(0.25) & 3.900(0.27) & 8.480(0.59) & 214.0(0.00)\\
    \cmidrule{2-8}
    \\[-0.8em]
    & & \multicolumn{6}{c}{($n = 200, p = 214$)}\\
    \\[-0.8em]
    & CA        & {\bf 0.756}(0.01) & {\bf 0.756}(0.01) & 0.736(0.01) & 0.754(0.01) & {\bf 0.756}(0.01) & 0.706(0.01) \\
    & NLL   & {\bf 290.8}(1.97) & 291.1(1.96) & 306.4(1.52) & 292.7(2.32) & 291.9(1.69) & 320.6(1.01)\\
    & Time      & 0.838(0.03) & 0.733(0.02) & 0.835(0.02) & 3.150(0.16) & 4.075(0.19) & 0.424(0.01)\\
    & No.V      & 3.020(0.15) & 3.060(0.15) & 4.740(0.25) & 4.760(0.32) & 8.980(0.59) & 214.0(0.00)\\
    \midrule
    \\[-0.8em]
    & & \multicolumn{6}{c}{($n = 100, p = 400$)}\\
    \\[-0.8em]
    \multirow{10}{*}{\textbf{Heart}}
    & CA        & 0.758(0.01) & 0.751(0.01) & 0.792(0.01) & 0.786(0.01) & 0.797(0.01) & \textbf{0.799}(0.01)\\
    & NLL   & 107.4(1.73) & 110.3(1.66) & 99.89(0.95) & 96.20(1.28) & {\bf 94.36}(0.93) & 103.2(1.03)\\
    & Time      & 3.943(0.09) & 3.623(0.11) & 0.237(0.01) & 1.208(0.05) & 1.483(0.06) & 0.383(0.01)\\
    & No.V      & 2.820(0.11) & 2.800(0.12) & 10.10(0.49) & 7.080(0.28) & 13.76(0.51) & 400.0(0.00)\\
    \cmidrule{2-8}
    \\[-0.8em]
    & & \multicolumn{6}{c}{($n = 200, p = 400$)}\\
    \\[-0.8em]
    & CA        & 0.798(0.01) & 0.788(0.01) & 0.813(0.01) & 0.802(0.01) & 0.805(0.01) & {\bf 0.828}(0.01)\\
    & NLL   & 47.86(0.95) & 49.53(0.94) & 46.31(0.41) & {\bf 44.90}(0.71) & 45.05(0.73) & 48.25(0.32)\\
    & Time      & 5.629(0.14) & 5.022(0.14) & 0.570(0.01) & 2.755(0.11) & 3.517(0.12) & 0.621(0.01)\\
    & No.V      & 5.520(0.15) & 5.480(0.14) & 14.28(0.67) & 9.340(0.37) & 17.38(0.58) & 400.0(0.00)\\
    \\[-0.8em]
    \bottomrule[1pt]
    \end{tabular}
    \label{tab:real_results_logistic}
    \vspace{-4mm}
\end{table*}

\section{Discussion}
In this paper we introduced a novel EM algorithm that allows us to efficiently find, for the first time, the exact, sparse posterior mode of the horseshoe estimator. The experimental results suggest that in comparison with state-of-the-art non-convex sparse estimators, the HS-EM estimator is quite robust to the underlying structure of the problem. Both the MCP and SCAD algorithm appear sensitive to correlation in the predictors, and the HS-like estimator, based on an approximation to the horseshoe prior, appears sensitive to the signal-to-noise ratio of the problem. In contrast, the HS-EM algorithm, even when not performing the best, appears to always remain competitive with the best performing method while selecting highly sparse models. Comparing the shrinkage profiles of the HS and HS-like estimator (see Figure~\ref{fig: shrinkage_profile}), the mixed performance of the HS-like procedure, is possibly due to the fact that HS-like approach more aggressively zeros out coefficients, making it more sensitive to misspecification of the global shrinkage parameter. We also conjecture that the conventional EM framework, in which the shrinkage hyperparameters $\bm{\lambda}$ are treated as missing data, provides a poorer basis for estimation of the global hyperparameter, in comparison to our approach of treating the coefficients as missing data, which potentially integrates the uncertainty of the parameters more effectively into the Q-function. This is a topic for further investigation.\\

Due to its prior preference for dense models, ridge regression can perform much worse than sparse alternatives if the underlying problem is sparse. However, it remained competitive or superior to other sparse methods on several of the real datasets. This strongly suggests that a method that is adaptive to the unknown degree of sparsity would be beneficial; we anticipate extending our algorithm to the generalised horseshoe discussed in Section~\ref{sec: Application to Horseshoe Estimator}, and automatically tuning our prior to produce a technique that will adapt to the sparsity of the problem.\\

While our HS-EM algorithm is able to locate, exactly, a sparse posterior mode of the horseshoe estimator, there is an obvious question as to what mode we are finding. Posterior modes (and means) are not invariant under reparameterisation, which suggests that maximising for $\lambda_j$ (with appropriate transformation of prior distribution) in place of $\lambda_j^2$, for example, will produce a new shrinkage procedure with potentially different properties. This is a focus of future research. The EM algorithm in this paper can also be easily extended to other popular priors such as the Bayesian lasso. Finally, given the relative simplicity of the M-step updates in this framework, it is of interest to consider extensions to non-linear models, such as neural networks. In this case, the required conditional expectations are unlikely to be available in closed form, but given access to a Gibbs sampler (or approximate Gibbs sampler) the procedure can easily be extended to utilise a standard stochastic EM implementation. This could potentially provide a simple, and (in expectation) exact, alternative to variational Bayes for these problems.


%
%
%
%

\newpage
\appendix

\section{Conditional Expectations for Normal Means Model} \label{apx: conditional expectation}
For the normal means model described in Section \ref{sec: Normal Means}, the conditional distribution of $\bm{\beta}$ is represented by a normal distribution with a mean of $(1-\kappa)y$ and variance of $\sigma^2(1-\kappa)$:
\begin{equation}\label{cond_post_mu}
    \bm{\beta}| \bm{\lambda, y},\tau \;\sim\; \rm{N} \left((1-\bm{\kappa})\bm{y}, \sigma^2 (1-\bm{\kappa})\right)
\end{equation}
where $\bm{\kappa} = 1/(1+\bm{\lambda^2}\tau^2)$. And to compute the expected value of $\bm{\beta}^2$, we will be taking the summation of the variance and the squared of the expected value of $\bm{\beta}$:
\begin{align}\label{expected_mu2}
\begin{split}
    E[\bm{\beta^2}] &= Var(\bm{\beta}) + E[\bm{\beta}]^2 \\
    &=\sigma^2 (1-\bm{\kappa}) + ((1-\bm{\kappa})\bm{y})^2
\end{split}
\end{align}

\section{Implementation of EM algorithm}
The proposed EM algorithm is summarised in Algorithm \ref{alg:EM}.
\begin{algorithm}
\caption{Basic EM algorithm applied to prior of choice}\label{alg:EM}
\begin{algorithmic}
\State {\bfseries Input:} Standardised data ${\bf X}$ and target ${\bf y}$
\State {\bfseries Initialise:} $\hat{\beta}_j^0 = {\left( ||{\bf x}_j||^{-2} ||{\bf x}_j||{\bf y} \right)}$, $\E{}{|| {\bf y}-{\bf X}\bm{\beta} ||^2} = 10^{10}$
\begin{itemize}
    \setlength\itemsep{1em}
    \item \textbf{Repeat}:
        \begin{itemize}
            \item (M-step) Update the shrinkage parameters $\bm{\lambda}^2$, $\tau^2$ and $\sigma^2$ by solving the optimisation problem detailed in Section 4.
            \item (E-step) Compute the conditional expectation of $\E{}{|| {\bf y}-{\bf X}\bm{\beta} ||^2}$ and $\E{}{\bm{\beta}^2}$ using either one of the approach describe in Section 3.2.
        \end{itemize}
        \vspace*{\baselineskip}
        Iterate until $\sum_{j=1}^p(|\hat{\beta}_j^{(t)} - \hat{\beta}_j^{(t+1)}|)/ (1+ \sum_{j=1}^p(|\hat{\beta}_j^{(t+1)}|)) < \omega$. Here we set the tolerance parameter, $\omega$ to $10^{-5}$
    \item If $|\hat{\beta}_j| < (5\sqrt{n})^{-1}$, set $\hat{\beta}_j = 0$
\end{itemize}
\end{algorithmic}
\end{algorithm}

\section{The use of Rue's algorithm for the E-step}

Given the conditional posterior distribution of the regression coefficients $\bm{\beta} \in \mathbb{R}^p$:
\begin{align}
\begin{split}\nonumber
    \bm{\beta}|\cdot \; &\sim \; N_p({\bf A}^{-1}{\bf X}^T{\bf y}, \sigma^2{\bf A}^{-1})\\
    {\bf A} \; &= \; ({\bf X}^T {\bf X} +  \tau^{-2} \bm{\Lambda}^{-1})
\end{split}
\end{align}
where ${\bm \Lambda} = \rm{diag}(\lambda^2_1, \cdots, \lambda^2_p)$. The basic idea for Rue's algorithm is to use the Cholesky decomposition of the covariance matrix ${\bf A^{-1}}$ and solve a series of linear systems. Therefore, to compute the posterior statistics from the above hierarchy:
\begin{enumerate}
    \item Compute the upper triangular Cholesky decomposition of the covariance matrix ${\bf A} = {\bf U}^T{\bf U}$.
    \item Compute $\E{}{\bm{\beta}} = {\bf U}\backslash\left[{\bf U}^T / \left({\bf X}^T{\bf y}\right)\right]$, where $a \backslash b$ denotes the back-solve operation and $a / b$ denotes the forward-solve.
    \item If using the exact expectation, ${\bf C} = {\bf U}^T / {\bf I}_p$ and ${\rm Var}[\beta_j] = \sigma^2||{\bf c}_j||^2_2$, otherwise, refer to the expressions for the approximate expectations detailed in Section 3.2.
\end{enumerate}
\section{Derivation of $\lambda^2$ updates for the horseshoe estimator}
\noindent \textbf{E-step}. Substitute the negative log-prior distribution for horseshoe given in Table 1 into Equation 5 and this give us:
\begin{align} 
&Q(\bm{\lambda}, \tau, \sigma^2|\hat{\bm{\lambda}}^{(t)}, \hat{\tau}^{(t)}, \hat{\sigma}^{2^{(t)}}) \nonumber \\
 &\; \; = \E{}{- \log p(\bm{\beta}, \bm{\lambda}, \tau, \sigma^2 \vbar {\bf y}) \: | \: \hat{\bm{\lambda}}^{(t)}, \hat{\tau}^{(t)}, \hat{\sigma}^{2^{(t)}}, {\bf y}} \nonumber \\
&\; \; = \left( \frac{n+p}{2} \right) \log \sigma^2 + \frac{\E{}{ ||{\bf y} - {\bf X}\bm{\beta}||^2 \:  | \: \hat{\bm{\lambda}}^{(t)}, \hat{\tau}^{(t)}, \hat{\sigma}^{2^{(t)}} }}{2 \sigma^2} + \frac{p}{2} \log \tau^2 \nonumber \\
& \; \; \; \; \; \; \; \; +\frac{1}{2}\sum_{j=1}^p \log \lambda_j^2 + \frac{1}{2 \sigma^2 \tau^2} \sum_{j=1}^p \frac{\E{}{\beta_j^2 \:  | \: \hat{\bm{\lambda}}^{(t)}, \hat{\tau}^{(t)}, \hat{\sigma}^{2^{(t)}} }}{\lambda_j^2} \nonumber \\
& \; \; \; \; \; \; \; \; + \sum_{j=1}^p \left[ \log(1+\lambda_j^2) + \frac{\log \lambda_j^2}{2} \right] + \log(1+\tau^2) \nonumber
\end{align}
\noindent \textbf{M-step}. To recover the $\lambda^2_j$ update in Equation 16, we minimize the above Q-function with respect to the local shrinkage hyperparameter:
\begin{align} \nonumber
    \hat{\lambda^2_j}^{(t+1)} &= \argmin{\lambda^2_j} \left\{ Q \left(\lambda^2_j \vbar \hat{\lambda^2_j}^{(t)} \right) \right\}\\
    &= \argmin{\lambda^2_j} \left\{\log \lambda_j^2 + \frac{W_j}{\lambda_j^2} + \log(1+\lambda_j^2)\right\} \label{eq:apx M-step}
\end{align}
where $W_j ={E[\beta_j^2]}/{(2\sigma^2\tau^2)}$. To solve for $\lambda^2_j$ that minimize function \ref{eq:apx M-step}, we first differentiate $Q \left(\lambda^2_j \vbar \hat{\lambda^2_j}^{(t)} \right)$ with respect to $\lambda_j$:
\begin{align}
    \frac{d \; Q \left(\lambda^2_j \vbar \hat{\lambda^2_j}^{(t)} \right)}{d\lambda^2_j} = \frac{1}{\lambda^2_j} + \frac{1}{1+\lambda^2} - \frac{W}{(\lambda^2)^2} \label{eq: derivative}
\end{align}
Then set derivative \ref{eq: derivative} to zero, solve for $\lambda^2_j$ and choose the positive solution, yielding:
\begin{align} \nonumber
    \hat{\lambda^2_j}^{(t+1)} &=  \frac{1}{4} \left( \sqrt{1+6W_j+W_j^2}+W_j-1  \right)
\end{align}
which recovers Equation 16 in Section 4.    

\end{document}